\pdfoutput=1
\documentclass{article}
\usepackage[utf8]{inputenc}
\usepackage{amsmath}
\usepackage{amssymb}
\usepackage{graphicx}
\usepackage{booktabs}
\usepackage[hidelinks]{hyperref}
\usepackage{cite}
\usepackage[margin=1in]{geometry}
\usepackage{array}
\usepackage{url}

\hypersetup{
  pdftitle={Truth for Believable AI: Expressed Doubt, Provenance, and Belief Revision as an Engineerable Stance},
  pdfauthor={Sebastian Cochinescu},
  pdfsubject={Epistemic expression and belief revision in conversational agents},
  pdfkeywords={language models, uncertainty expression, belief revision, provenance, conversational agents}
}

\newcommand{\PAR}{\textsc{par}}
\newcommand{\RET}{\textsc{ret}}
\newcommand{\INF}{\textsc{inf}}
\newcommand{\TOLD}{\textsc{told}}
\newcommand{\NK}{\textsc{not-knowing}}
\newcommand{\UN}{\textsc{unsure}}
\newcommand{\CF}{\textsc{confident}}
\newcommand{\WR}{\textsc{wrong-revising}}

\title{Truth for Believable AI: Expressed Doubt, Provenance, and Belief
Revision as an Engineerable Stance}

\author{
Sebastian Cochinescu\\
\textit{University of Bucharest}\\
\textit{Email: sebastian.cochinescu@drd.unibuc.ro}
}

\date{September 2026}

\begin{document}

\maketitle

\begin{abstract}
Conversational agents often express answers in a uniformly confident
register. We test whether expressed uncertainty, provenance-aware assertion,
and explicit belief revision can be implemented as a behavior layer over a
fixed language model; we do not test believability or trust. The layer combines
three epistemic states, per-claim confidence and typed provenance, a
provenance-gated expression rule, and a persistent revision store with
auditable acknowledgments and partial resistance to false corrections. We
evaluate it on a constructed, mechanically scored multi-session benchmark
using a synthetic model and Qwen2.5-0.5B-Instruct. The synthetic instrument
passes all five checks. On the real model, acknowledgment soundness, a
by-construction guarantee, holds in 100\% of cases, and true corrections are
accepted more often than false ones (0.44 vs.\ 0.15 on held beliefs; 0.875
vs.\ 0.420 including rule-accepted corrections of unheld facts), but the
pre-specified expression-fidelity, contradiction-separation, and provenance
margins fail. A disclosed post hoc analysis shows that expression gated on
mean answer-token probability ranks correctness below chance end to end (AUC
0.41, conversation-clustered), whereas gating on sampling consistency discriminates (AUC 0.66). A
consistency-gated configuration selected from this finding and evaluated under
a separately committed protocol meets the conversation-level manipulation and
capability-equivalence criteria and replicates on a redrawn conversation set.
The manipulation result is selection-dependent, and both criteria remain
unresolved when uncertainty is clustered over the 60 facts. The supported
conclusions are limited to the by-construction audit guarantee,
store-dependent partial correction discrimination, and a benchmark- and
model-specific failure of token-probability gating; scaling the fact base is
required before human evaluation.
\end{abstract}

\section{Introduction}
\label{sec:problem}

A large language model answers a question it cannot answer in the same fluent,
unhedged register it uses for one it can. The confident hallucination is the
canonical failure, but the deeper problem is the \emph{uniformity}: nothing in
the surface behavior distinguishes knowledge from guess, retrieval from
inference, or a fact the user supplied five turns ago from one the model was
trained on. Kalai et al.\ argue that this follows from the evaluation regime: benchmarks
that score accuracy alone make confident guessing the optimal policy and
penalize ``I don't know''~\cite{kalai2026evaluating}. The estimation side of
the problem is well studied in laboratory settings: models carry
usable self-knowledge~\cite{kadavath2022language}, semantic entropy detects
confabulation~\cite{farquhar2024detecting,kuhn2023semantic}, and models can be
taught to verbalize numeric confidence~\cite{lin2022teaching,xiong2024can}.
What is missing is the \emph{behavior layer}: the machinery that turns an
epistemic signal into graded assertion, explicit declining, visible
self-correction, and source-aware caution, and that keeps those behaviors
consistent across sessions.

This paper builds and measures that layer. The motivation is a believability
hypothesis inherited from a companion framework
paper~\cite{cochinescu2026perceivedagi} and consistent with human-subject
findings that uncertainty expression changes reliance and
trust~\cite{kim2024imnotsure,xu2025confronting}: an agent that can say ``I am
not sure,'' show where a claim came from, and visibly revise a stated belief
reads as more trustworthy and more mind-like than one that cannot. This study
reports no human-subjects data and makes no claim about trust, believability,
or perceived mind; those hypotheses are deferred to a preregistered perception
study (Section~\ref{sec:handoff}). We present a formal model
(Section~\ref{sec:model}), an implementation over a fixed base model
(Section~\ref{sec:impl}), a constructed contradiction benchmark, and a staged,
pre-specified evaluation (Sections~\ref{sec:protocol}%
--\ref{sec:results}). The evidence class is synthetic, seeded, single-machine
measurement, mechanically scored against generator-injected ground truth. The
closest prior work, linguistic calibration of a dialogue
agent~\cite{mielke2022reducing}, maps a scalar correctness estimate to hedging
words in single turns. Our contribution is the combination of a three-state
expression policy (discrete confidence levels themselves already appear
there), a typed provenance gate, a persistent store with a defined revision
operator, and auditable acknowledgment behavior, each varied by a dedicated
ablation arm.

We report all pre-specified outcomes, including three margins that failed on
the real model and a disclosed post hoc protocol amendment whose amended check
also failed. The failures expose a
conditional degeneracy in anchor-based expression-fidelity scoring, a scoping
limit of contradiction-based consistency metrics under deterministic decoding,
and an extractor-dependence result in which the two confidence sources yield
oppositely signed end-to-end discrimination. A final pass on the configuration identified by that result was governed by a
separately committed protocol and passes its specified checks under the
conversation-level analysis, with a disclosed fact-level sensitivity caveat
(Section~\ref{sec:handoff}). Each of the layer's four mechanisms is varied by
a dedicated ablation arm in Section~\ref{sec:protocol}
(Table~\ref{tab:diff} positions them against related work). The arm contrasts
are single realizations with independent retrieval-sampling streams, so they
describe each mechanism's contribution rather than causally isolate it
(Section~\ref{sec:results}).

\section{The truth stance and its measurement}
\label{sec:stance}

The companion framework paper defines the \emph{truth stance} behaviorally:
the disposition to expose epistemic state (to say ``I am not sure,'' to admit
a mistake, to revise a stated belief), explicitly \emph{not} statistical
calibration or answer accuracy, which are competencies. This distinction
matters because the stance must be manipulable with capability held constant:
a fixed-capability model can exhibit more or less of it, which is what makes
an ablation over the same base model meaningful.

This paper measures calibration-like quantities, so the relation between the
stance and its manipulation check must be explicit. The stance is the
disposition to expose epistemic state. \emph{Expression fidelity}, whether
expressed-confidence categories track empirical accuracy, is the
\emph{manipulation check} that distinguishes evidence-sensitive expression
from performative humility. The framework itself requires this check:
expressed doubt must be congruent with the available evidence, not blanket
hedging. We operationalize evidence-congruence as expression-level expected
calibration error (expression-ECE), an operationalization that
Section~\ref{sec:results} amends, with disclosure, after its degeneracy is
measured. We flag the choice explicitly: congruence with evidence and
accuracy tracking are close but not identical notions, and the always-hedged
control arm exists to separate evidence-sensitive doubt from indiscriminate
doubt. That arm is the discriminant-validity control. It maximizes expressed
doubt while its expression-discrimination stays at chance by construction
(Section~\ref{sec:results}), so the manipulation check rewards
evidence-sensitivity rather than doubt volume. The check is end to end,
jointly determined by the confidence signal and the expression policy, so a
failed check localizes to that composite and not to the stance disposition
alone; the extractor-dependence result in Section~\ref{sec:results} is such
a signal-side failure. We also note a design tension visible before any data.
The gating table caps inferred and told content below \textsc{assert}
whatever its confidence (source-caution by design; retrieved content asserts
in the high band), which guarantees underconfident bins under fixed anchors,
so an anchor-ECE operationalization is in structural tension with a
provenance-capping layer. We retained the pre-specified check as frozen
rather than redesigning around the recorded tension, and report its outcome;
the tension resurfaces, measured, in Sections~\ref{sec:results}
and~\ref{sec:handoff}. We calibrate the \emph{expression channel over a fixed
base model}; we neither improve nor claim to improve the model's underlying
calibration, and we never define the stance as the competency.

\section{A formal expression-and-revision layer}
\label{sec:model}

\paragraph{Claims and epistemic state.}
The unit is a \emph{claim} $c = (s, a, v)$: subject, attribute, value. Each
claim the layer is about to express carries an epistemic state
$\sigma(c) = (\kappa(c), \pi(c))$ where $\kappa(c) \in [0,1]^{E}$ is a vector
of confidences from $E \ge 2$ distinct extractors (separately computed but not
statistically independent, since both consult the same base model) and
$\pi(c) \in \{\PAR, \RET, \INF, \TOLD\}$ is a \emph{provenance tag}. The
combined confidence is the conservative rule
$\hat\kappa(c) = \min_e \kappa_e(c)$: no single inflated extractor can unlock
a strong assertion. The rule was chosen on this safety argument alone, before any data. Its
cost, that a single noisy extractor can force unnecessary hedging, is
accepted by design, and the per-extractor runs in Section~\ref{sec:results}
measure what each extractor contributes alone. Provenance is assigned by
pipeline instrumentation, by \emph{which resolution path produced the claim}
(model parameters, retrieval store, derivation rule, or a user statement this
session), never by a classifier. The taxonomy is an operational definition, not an epistemological
claim, and the boundary between parametric and inferred content is ambiguous.

\paragraph{Three behavioral states.}
The layer distinguishes three expressible epistemic conditions:
\NK{} (no candidate value, or $\hat\kappa < \theta_{\mathrm{floor}}$),
\UN{} ($\hat\kappa < \theta_{\mathrm{high}}$), and \CF{} (otherwise), with
$\theta_{\mathrm{floor}} = 0.15$ and $\theta_{\mathrm{high}} = 0.75$
specified before evaluation. A fourth state, \WR{}, is \emph{not} reachable from confidence: it is entered
only through the revision operator's defined triggers below. In the released
implementation it is a logged condition (a revision-log entry and, on accepted
corrections to a held value, the acknowledgment utterance) rather than a value
the state assigner returns. The three-state model rests on the claim that
not-knowing, being unsure, and being wrong are behaviorally distinct
conditions with distinct correct behaviors (decline, hedge,
acknowledge-and-revise), which a single confidence threshold conflates.
Whether the distinction buys anything measurable is what the threshold-only
ablation tests.

\paragraph{Expression function.}
Expression categories are ordered
$\textsc{decline} < \textsc{hedge-low} < \textsc{hedge-high} <
\textsc{assert}$, each with a fixed surface template and, for scoring, a
nominal confidence anchor (0.3, 0.6, 0.9; representative band values fixed
at freeze, not fitted). After the degeneracy finding of
Section~\ref{sec:results}, anchor-ECE functions as a stress test rather than
the primary operationalization; \textsc{decline} carries no asserted content
and is scored as coverage, not fidelity. The expression function applies a
frozen \emph{gating table} $G(\pi, \mathrm{band}(\hat\kappa))$
(Table~\ref{tab:gating}) with band edges at $0.40$ and $0.75$. The table
encodes three design rules, recorded before evaluation: inferred claims never
reach bare \textsc{assert}; told-this-session claims are never asserted as the
agent's own knowledge (the implemented templates hedge them as ``I believe
$v$'' without naming the user as the source; explicit source attribution is a
template extension, not an evaluated behavior); and low-confidence parametric
content declines rather than hedges. The no-provenance ablation
replaces $G$ with the row-independent confidence map (bottom row), so the
difference between the two arms is the contrast for what provenance typing
adds (one realization per arm; Section~\ref{sec:results}).

\begin{table}[t]
\centering
\caption{The pre-specified provenance-gating table $G$: provenance class $\times$
combined-confidence band $\rightarrow$ expression category. The bottom row is
the no-provenance ablation's row-independent map. Recorded before the
prototype was built.}
\label{tab:gating}
\small
\begin{tabular}{lccc}
\toprule
Provenance & band \textsc{low} ($<0.40$) & \textsc{mid} & \textsc{high} ($\ge 0.75$)\\
\midrule
\PAR{} (parametric) & \textsc{decline} & \textsc{hedge-low} & \textsc{assert}\\
\RET{} (retrieved) & \textsc{hedge-low} & \textsc{hedge-high} & \textsc{assert}\\
\INF{} (inferred) & \textsc{decline} & \textsc{hedge-low} & \textsc{hedge-high}\\
\TOLD{} (told) & \textsc{hedge-low} & \textsc{hedge-high} & \textsc{hedge-high}\\
\midrule
\emph{ablation (no provenance)} & \textsc{hedge-low} & \textsc{hedge-high} & \textsc{assert}\\
\bottomrule
\end{tabular}
\end{table}

\paragraph{Belief store and revision operator.}
Beliefs persist in a session-spanning store $B$ of claims with their states.
The \WR{} state is entered by exactly three triggers: (i) an \emph{accepted
user correction}; (ii) an \emph{insertion conflict} (a new claim contradicts
a stored one; the higher combined confidence wins); (iii) a \emph{retrieval
conflict} (a retrieval result contradicts a stored belief). The
correction-acceptance rule is stated formally because an epistemic-humility
layer that accepts every correction is sycophancy~\cite{sharma2023sycophancy},
not humility. A correction of stored belief $b$ is accepted iff
$\hat\kappa(b) < \theta_{\mathrm{acc}}$ \emph{or} the correction is
corroborated by the retrieval store ($\theta_{\mathrm{acc}} = 0.75$, specified
before evaluation); otherwise the belief is kept and the resistance is logged
(the released prototype records the refusal but renders no disagreement
utterance). A correction concerning a claim absent from the store is accepted
unconditionally, logged, and not acknowledged, since there is no prior value
to retract; this case is frequent on the benchmark, and the scored acceptance
rates include it (Section~\ref{sec:results}). One edge of this rule is a
deliberate scope choice we disclose: for weakly held beliefs ($\hat\kappa <
\theta_{\mathrm{acc}}$) a user correction is accepted even when the retrieval
store contradicts it, so retrieval corroboration can rescue a strongly held
belief but never vetoes a correction to a weak one. The residual threat is
explicit: an adversarial user can rewrite any weakly held belief without
corroboration. Adjudicating such contested corrections is future work; on
this benchmark the acceptance statistics below measure the rule as stated.
Retrieval conflicts apply the same weak-belief threshold with no
corroboration clause: a stored belief with $\hat\kappa(b) <
\theta_{\mathrm{acc}}$ yields to the retrieved value, a stronger one resists.
Accepted revisions re-enter the store with fixed post-revision states (an
accepted correction as (\TOLD{}, confidence $0.85$), a retrieval-conflict
winner as (\RET{}, $0.80$)), and a correction that agrees with the stored
value is a no-op and creates no revision-log entry. Every revision, accepted or resisted, appends to a
revision log, and the layer's acknowledgment utterance (``I was wrong about
$X$---I said $u$, it is $v$'') is emitted only when a matching accepted log
entry exists. Auditability here is a \emph{soundness} guarantee: every public
admission of error traces to a recorded state change. The guarantee is
checked on the event linkage (an acknowledged turn must carry the key of an
accepted log entry). It certifies neither the wording of the admission (the
template's ``I said $u$'' names the stored prior value, which the layer need
not have uttered earlier in the conversation) nor the truth of the accepted
value. The reciprocal direction is a measured quantity, not a guarantee: what
fraction of accepted state changes are flagged as acknowledged, and what
fraction of those flags render the admission utterance. The flag is set on
accepted corrections to a held value, which render the utterance, and on
retrieval or insertion conflicts detected at response time, which are flagged
and logged but answered with an ordinary expression. Accepted corrections of
unheld claims and silent store-maintenance updates change the store and the
log directly without passing through the uttered \WR{} behavior at all; \WR{}
names the \emph{expressed} condition, not every logged state change.
Section~\ref{sec:results} reports both directions. Corroboration consults
only the retrieval store, never ground truth: the layer cannot peek at the
answer key.

\paragraph{Self-model term.}
The layer can attach uncertainty to reports about its own store (``my memory
of this may be stale''), and we include the term as a \emph{design element
only}, bounded by the role-play frame~\cite{shanahan2023roleplay}:
introspective reports are generated behavior, not privileged access, and the
live controversy over machine theory of
mind~\cite{strachan2024testing,kosinski2024evaluating,ullman2023trivial} is
kept out of our claims, since our states are computed pipeline quantities,
not introspection. No numbered claim rests on the self-model term.

\paragraph{Worked example.}
A five-turn exchange, mechanically checkable (its store transitions are
covered by a unit test): the
user asks a subject's attribute; the layer resolves it parametrically at
$\hat\kappa = 0.9$ (\CF{}, \PAR{}) and asserts, storing the belief. The user
then corrects it without corroboration; since $\hat\kappa = 0.9 \ge
\theta_{\mathrm{acc}}$, the correction is \emph{resisted} and logged. A
corroborated correction next turn is accepted: the store updates (provenance
\TOLD{}), the acknowledgment fires, and the log gains its accepted entry. A
re-ask now yields the revised value; the scorer does not count this as a
contradiction because the change follows an accepted revision-log entry. Finally a question with no candidate resolves to \NK{} and
is declined.

\section{Related work}
\label{sec:related}

Prior work addresses the components of this layer separately; none, to our
knowledge, combines them into a behavioral epistemic interface with auditable
correction acknowledgment and ablation-compared mechanisms, and each gap
below corresponds to one of our ablation arms.

\emph{Estimation} is imported, not contributed:
models know much of what they know~\cite{kadavath2022language}, semantic
uncertainty, entropy, and sampling consistency supply usable
signals~\cite{kuhn2023semantic,farquhar2024detecting,manakul2023selfcheckgpt},
and elicited or verbalized confidence works well enough to build
on~\cite{lin2022teaching,xiong2024can,tian2023just}.
The closest related work concerns \emph{expression}: linguistic
calibration trains a dialogue agent so hedging words track a scalar
correctness estimate~\cite{mielke2022reducing}, extended to long-form
generation by decision-calibration objectives~\cite{band2024linguistic}, with
epistemic markers studied for their downstream
effects~\cite{zhou2023navigating}. This work calibrates expression in single turns (Mielke et al.\ already
control three linguistic-confidence levels, don't-know, low, and high, with
control tokens) but has no persistence, no provenance typing, and no
revision; the discrete levels are not our contribution, their combination
with the rest of the layer is.

\emph{Abstention} supplies the nearest neighbor to
our \NK{} state: unanswerable-question benchmarks~\cite{rajpurkar2018know},
selective prediction~\cite{kamath2020selective}, refusal-aware
tuning~\cite{zhang2024rtuning}, honesty as an alignment
objective~\cite{yang2023alignment}, sycophancy under user pushback as the
failure mode our acceptance rule resists~\cite{sharma2023sycophancy}, and a
recent survey~\cite{wen2025know}
mostly treat abstention as an answer-or-refuse boundary, static and
single-turn (the survey itself formalizes partial abstention and dynamic
information acquisition), whereas in our layer abstention is one endpoint of
a graded, provenance-typed policy over persistent state. \emph{Attribution}
measures whether output is
supported by sources~\cite{rashkin2023measuring,bohnet2022attributed}, makes
models emit citations~\cite{gao2023enabling}, repairs unattributed claims
post hoc~\cite{gao2023rarr}, or gates generation on retrieval support with
reflection tokens~\cite{asai2024selfrag}; this is provenance that annotates
or gates generation, with no graded cap on assertion strength.

\emph{Belief revision} provides a theoretical context and a contrast: the AGM
operators~\cite{alchourron1985logic} locate our store-level
revision in a formal tradition (cited as lineage, not implemented logic);
weight-level knowledge editing~\cite{meng2022rome,meng2023memit,
yao2023editing} is the contrast case (a different substrate, parametric
memory, rather than a competing design), and recent conceptual work argues
that editing lacks rational belief-revision
foundations~\cite{hase2024fundamental}, an argument that motivates an
external, auditable store such as ours.
Behavioral belief-revision evaluation exists for reasoning~\cite{wilie2024belief}
and belief-like agent memory is emerging~\cite{liao2026beliefmemory}, but for
internal correctness, not user-facing epistemic communication.

\emph{Consistency benchmarks} detect contradictions in
dialogue~\cite{welleck2019dialogue,nie2021decode} and probe long-horizon
memory~\cite{xu2022goldfish,maharana2024locomo}; ours differs by
\emph{constructing} contradictions and corrections so that revision behavior,
not only consistency, is mechanically scorable. Finally, hedging itself is an
epistemic semantic device~\cite{lakoff1973hedges,hyland1998hedging} rather
than a politeness strategy~\cite{brown1987politeness}; our expression
function encodes evidence state, not face-saving. Table~\ref{tab:diff}
summarizes these comparisons.

\begin{table}[t]
\centering
\caption{What each line of related work contributes and lacks.
$\bullet$ = present, $\circ$ = partial/single-turn (discrete confidence
levels in linguistic calibration, partial abstention in the abstention
literature), --- = absent. The table positions related work; it does not map
mechanisms to ablation arms.}
\label{tab:diff}
\small
\begin{tabular}{lcccc}
\toprule
 & multi-state & provenance & persistent store & auditable\\
 & expression & gating & + revision & acknowledgment\\
\midrule
Linguistic calibration~\cite{mielke2022reducing,band2024linguistic} & $\circ$ & --- & --- & ---\\
Abstention / refusal~\cite{zhang2024rtuning,kamath2020selective,wen2025know} & $\circ$ & --- & --- & ---\\
Attribution / citation~\cite{rashkin2023measuring,gao2023enabling,asai2024selfrag} & --- & $\circ$ & --- & ---\\
Knowledge editing~\cite{meng2022rome,meng2023memit} & --- & --- & $\circ$ & ---\\
Belief-like memory~\cite{liao2026beliefmemory,wilie2024belief} & --- & --- & $\bullet$ & ---\\
\textbf{This layer} & $\bullet$ & $\bullet$ & $\bullet$ & $\bullet$\\
\bottomrule
\end{tabular}
\end{table}

\section{Implementation}
\label{sec:impl}

The layer is a thin instrument over a fixed base model: extractor adapters, the
expression mapper with the frozen gating table, the belief store with its
revision log, and a provenance tagger that records the resolution path. Answer
resolution order is fixed: store hit, then parametric query, then retrieval
lookup, then one-step derivation via a public rule table; the path taken
\emph{is} the provenance tag. Greedy decoding was frozen for determinism and reproducibility before its
dampening effect on the contradiction metric was known (a scoping consequence
Section~\ref{sec:results} reports). Two extractors run throughout: a
signal-style extractor (latent reliability in the synthetic stage; in the
real-model stage, the mean softmax probability of the generated answer's
tokens up to and including end-of-sequence, unnormalized for length and
computed on the greedy pass, so no sampling-temperature artifact enters it)
and a consistency
extractor
in the SelfCheckGPT tradition~\cite{manakul2023selfcheckgpt}: agreement of $k$
answers resampled at temperature $1.0$ (top-$p$ $0.95$) with the greedy
answer, using the same procedure in both stages; exact answer normalization
(lowercasing, de-accenting, snap-to-known-value by containment) and agreement
rules are included in the companion repository. Retrieval draws
on a small document store
(generator coverage probability 0.5 and stale probability 0.1; the realized
store shared by every grid holds 34 of the 60 subjects, one of them stale) so
that retrieval conflicts occur by design.

\begin{figure}[t]
\centering
\includegraphics[width=0.55\linewidth]{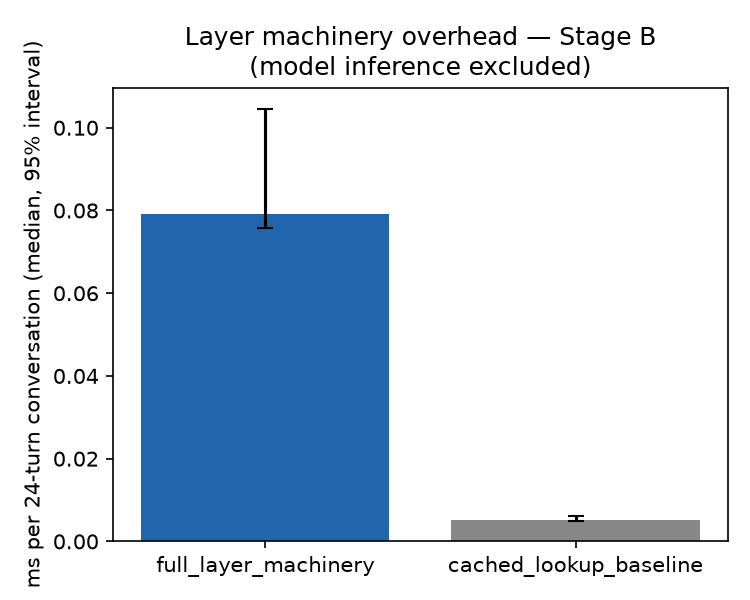}
\caption{Stage-B layer overhead: the expression-and-revision machinery runs
at a median 0.08\,ms per 24-turn conversation, against 0.005\,ms for a bare
cached lookup (model inference, measured separately at a median 107\,ms per
query, is excluded by caching). Bars show medians with the 2.5--97.5th
percentile spread of repeated runs, not bootstrap intervals. In this prototype
the layer's cost is negligible relative to measured model inference.}
\label{fig:overhead}
\end{figure}

Everything is seeded and reproducible from one command per stage. Constants
are recorded in a version-controlled protocol with a frozen/provisional status
legend, and benchmark artifacts are protected by SHA-256 manifests; the
reproduction script refuses to run if a frozen file's hash has drifted. The
prototype is an instrument, not a product: no UI, no serving stack. The added
machinery runs at a median 0.08\,ms per 24-turn conversation, against
0.005\,ms for a bare cached lookup (Figure~\ref{fig:overhead}); this is
orchestration overhead measured over cached model outputs on one machine.
End-to-end cost is dominated by model calls instead:
consistency extraction spends $k{=}8$ sampled generations per new fact on
top of the greedy pass (that pass alone had a median latency of 107\,ms per
query on our hardware, a macOS machine using the MPS backend), amortized here
by the per-fact cache.

\section{Evaluation protocol}
\label{sec:protocol}

The protocol was maintained in the project repository, not deposited in an
independent registry. For Stage~B, the protocol freeze and results first appear
in the same repository commit, so their ordering rests on the author's
working-tree record rather than an independently timestamped registration. We
therefore describe the Stage-B tests as \emph{pre-specified}, not
preregistered. The Stage-C configuration and its robustness draw were each
committed separately before execution, with prior knowledge recorded in the
protocol. This distinction does not alter any threshold or result.

\paragraph{Staged evaluation.}
Stage~A validates the \emph{instrument} against an LLM-free synthetic base
model: a seeded world of 60 subjects with two attributes plus one derived
attribute, and a base-model stub holding a deliberately corrupted copy
(correct/wrong/uncertain/absent facts with latent reliabilities, plus a
confident-hallucination path). Ground truth is knowable by construction, so
every metric is mechanical and directional expectations can be asserted as
tests. Stage~B is the pre-specified grid on a pinned real model
(Qwen2.5-0.5B-Instruct, revision \texttt{7ae557604adf67be50417f59c2c2f167def9a775},
greedy answers, $k{=}8$ consistency samples), over a
real 60-fact geography world (country $\rightarrow$ capital parametric;
continent derived via a public capital-to-continent rule; ambiguous capitals
excluded by construction, under an administrative-capital convention whose one
remaining edge is Eswatini, entered as Mbabane although Lobamba is the
legislative seat). The real model's error pattern is its own; there is no injected corruption.
All margins below were recorded by the author before the Stage-B grid ran and were
set by design judgment
as the smallest differences we considered practically meaningful; no
prospective power analysis informed them, and Section~\ref{sec:results}
reports achieved precision instead.

\paragraph{Arms.}
Seven configurations compare mechanism settings using the same base model:
\emph{full} (three states, gating table, store, acknowledgment);
\emph{uniform} (always assert; the truth-ablated surface);
\emph{always-hedged} (performative humility: uniform doubt regardless of
evidence); \emph{threshold-only} (confidence map, no three-state floor, no
provenance); \emph{store-without-acknowledgment} (revises silently);
\emph{no-provenance} (three states, confidence map); \emph{stateless} (no
store).

\paragraph{Benchmark.}
120 seeded three-session conversations (24 turns each) over the world: asks,
re-asks (at least one across a session boundary), user statements, and user
corrections, with half of the statements and corrections false by injection.
Labels are the generator's injection record, never any arm's output. The
benchmark ships versioned (instances, labels, manifest with hashes and
generation parameters) with a language-agnostic scorer, so any implementation
of an expression layer can be scored against it.

\paragraph{Metrics and pre-specified margins.}
C2 (expression fidelity): expression-ECE against the category anchors, unit =
conversation, cluster-bootstrap CIs (percentile, 2{,}000 seeded resamples of
whole conversations, the construction used for every interval in this
paper except the timing spreads of Figure~\ref{fig:overhead}, which are
run-time percentiles, and the intervals explicitly labeled fact-clustered,
which resample subjects instead); margin: full must undercut both the
uniform and always-hedged controls by $\ge 0.02$ under both extractors.
C3 (revision): unacknowledged contradiction rate on re-asks (an accepted
logged revision legitimizes a change); margin: stateless exceeds every store
arm by $\ge 0.03$; acknowledgment auditability must be 100\%; false-correction
acceptance must undercut true-correction acceptance by $\ge 0.10$.
C4 (provenance): confidently-asserted-false rate; margin: no-provenance
exceeds full by $\ge 0.02$, with the coverage cost reported alongside.
Capability equivalence: paired cluster-bootstrap CI of the
answer-when-given-accuracy difference (full $-$ uniform) within $\pm
0.05$, in effect answer-conditional accuracy equivalence at each arm's
achieved coverage, with coverage reported alongside.
The downstream selection rule (Section~\ref{sec:handoff}) requires the C2 margins
\emph{and} capability equivalence; a null result does not satisfy it, and
nothing is relabeled after the fact. One protocol amendment was added after the
Stage-B grid ran and is disclosed as post hoc where its results appear
(Section~\ref{sec:results}): the original verdicts are never displaced by it.
A third grid, Stage~C, evaluates the configuration that Stage~B's mechanism finding identifies (the
layer gated on the consistency extractor alone, everything else held
identical) under its own freeze, committed to the
repository \emph{before} that grid ran, with an explicit register of which
values were already known from Stage~B's robustness runs and which were not.

\section{Results}
\label{sec:results}

\subsection{Stage A: the instrument behaves as designed}

All five pre-specified instrument checks pass on the synthetic model. The
full layer's expression-ECE (0.161 combined; 0.151/0.199 per extractor) beats
uniform (0.237) and always-hedged (0.368) under both extractors; the stateless
arm self-contradicts (0.065) where every store arm is at zero; provenance
gating cuts the confidently-false rate from 0.141 (no-provenance) and 0.315
(uniform) to 0.094, with the coverage cost visible (0.838 vs.\ 0.932);
acknowledgment auditability is 73/73; and the acceptance rule discriminates
true from false corrections (0.77 vs.\ 0.54 accepted). Stage~A numbers
validate the instrument and are not the paper's headline results.

\subsection{Stage B: pre-specified verdicts on a real model}

\begin{figure}[t]
\centering
\includegraphics[width=\linewidth]{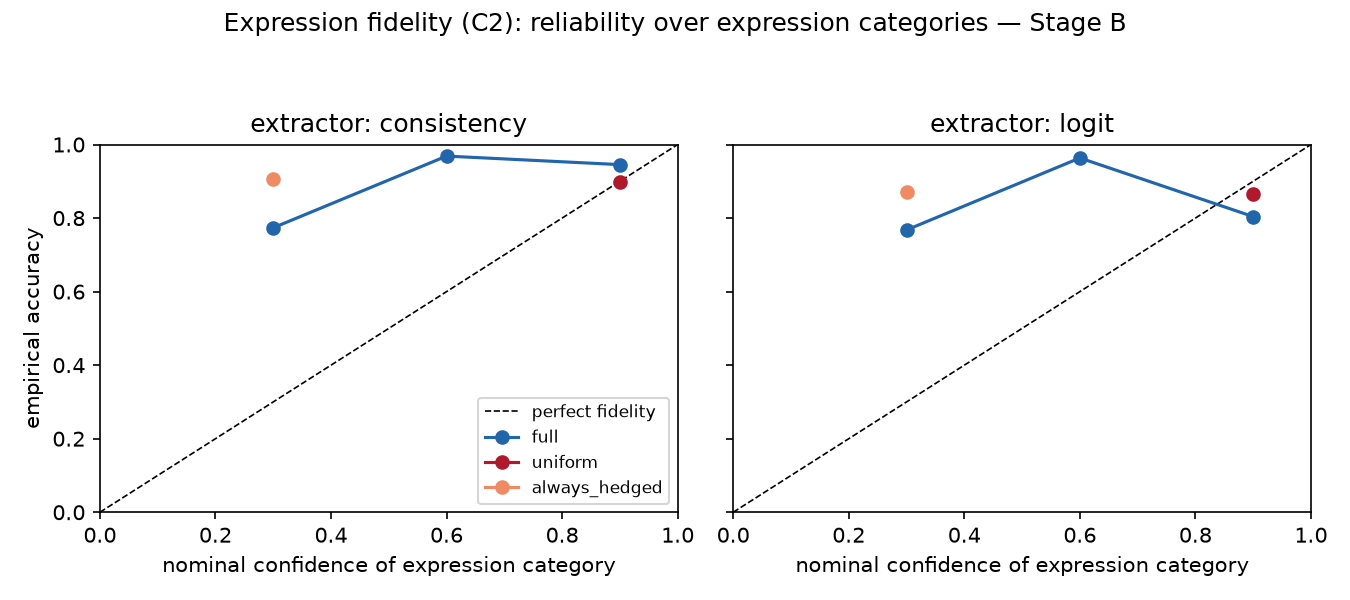}
\caption{Stage-B reliability over expression categories, per extractor. The
full layer (blue) spreads content across bins whose accuracy is not monotone in category
under either extractor (consistency: 0.77, 0.97, 0.95; logit: 0.77, 0.96,
0.80); overall expression discrimination is nonetheless positive under
consistency gating and negative under logit gating (the AUC analysis, with
its clustering caveat); the uniform
control is a single bin whose accuracy happens to sit at the 0.9 anchor, the
anchor-degeneracy that breaks the pre-specified ECE margin.}
\label{fig:reliability}
\end{figure}

\begin{figure}[t]
\centering
\includegraphics[width=0.72\linewidth]{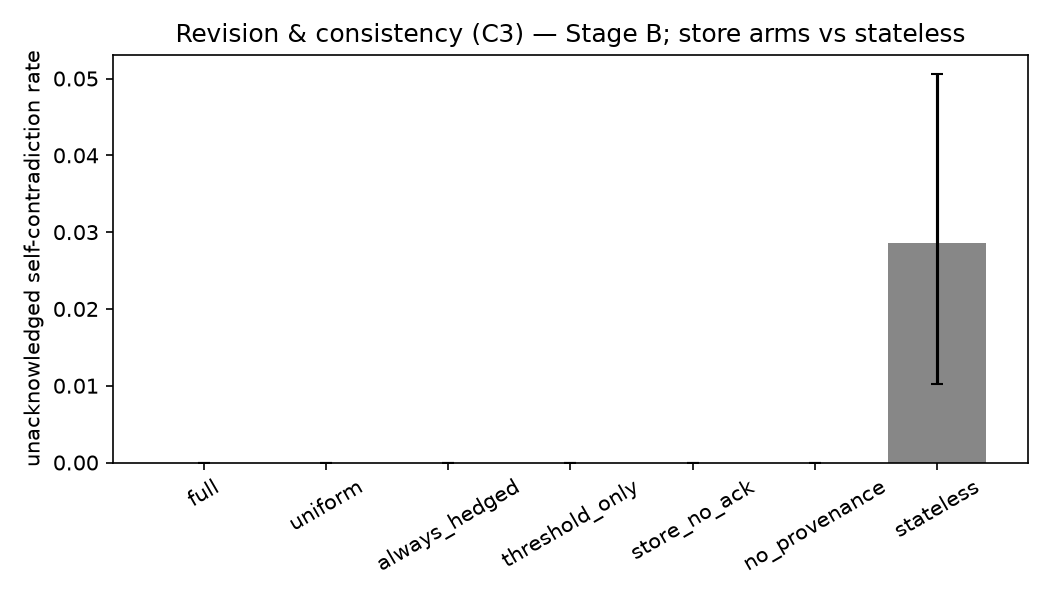}
\caption{Stage-B unacknowledged self-contradiction rate by arm. Store arms
are at zero; the stateless arm's 0.029 falls just short of the 0.03 margin
because greedy per-fact decoding makes the base model deterministic. Error
bars are 95\% conversation-cluster-bootstrap intervals over the re-ask
opportunities.}
\label{fig:consistency}
\end{figure}

\begin{figure}[t]
\centering
\includegraphics[width=\linewidth]{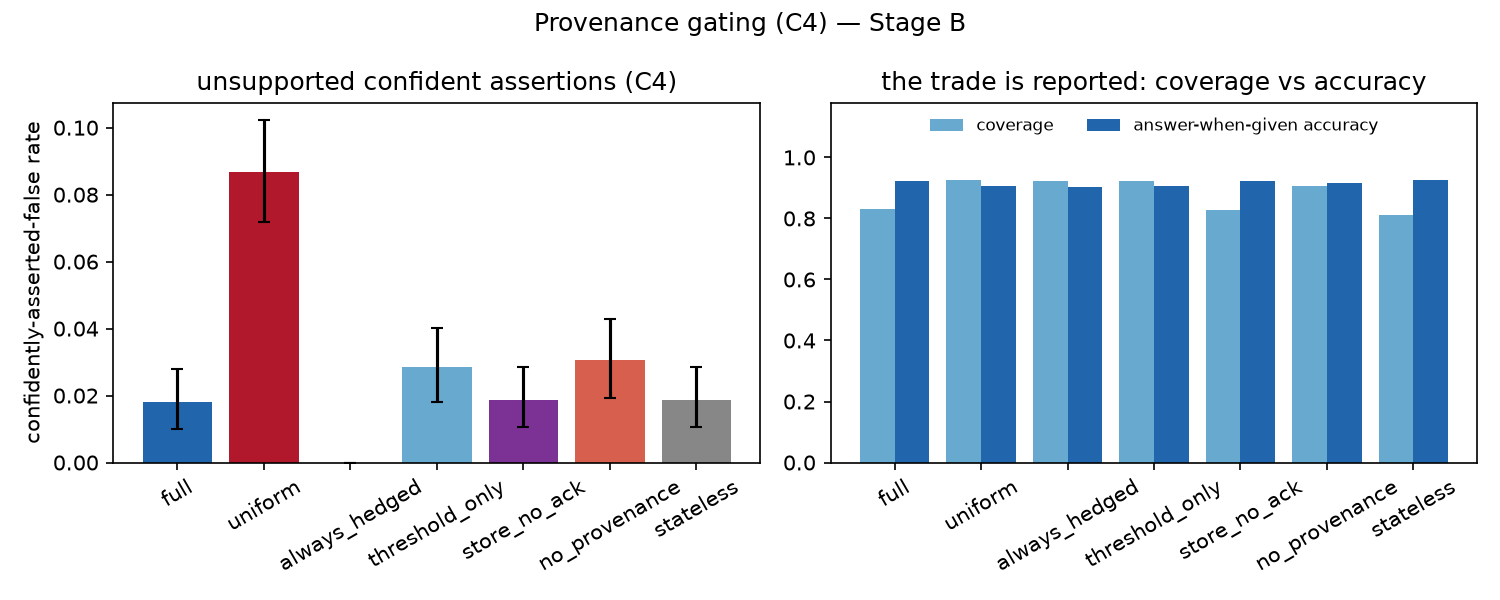}
\caption{Stage-B confidently-asserted-false rates (left; with 95\%
cluster-bootstrap intervals) and the coverage/accuracy trade (right). Gating
cuts confident falsehoods roughly five-fold versus uniform assertion; the
cost is paid in coverage, not accuracy.}
\label{fig:assertions}
\end{figure}

The pinned 0.5B model answers 45/60 capitals correctly, produces a
refusal-style non-answer for exactly one of the 60 queries (the pipeline
ingests it as an ordinary low-consistency value), and hallucinates fluently
with high signal confidence (e.g.\ a superseded capital asserted at 0.94 mean
token probability), while the consistency extractor's raw agreement spreads
from 0.00 on the model's worst guesses to 1.00 on well-known facts. Table~\ref{tab:verdicts} gives the seven frozen verdicts;
Figures~\ref{fig:reliability}--\ref{fig:assertions} show the underlying
distributions.

\begin{table}[t]
\centering
\caption{Stage-B pre-specified margin verdicts (margins author-recorded before
the grid ran). All numbers trace to the accompanying results files.}
\label{tab:verdicts}
\small
\begin{tabular}{lll}
\toprule
Pre-specified check & Margin & Verdict\\
\midrule
C2 fidelity, logit extractor & full ECE $\le$ controls $-$ 0.02 & \textbf{fail}\\
C2 fidelity, consistency extractor & full ECE $\le$ controls $-$ 0.02 & \textbf{fail}\\
C3 contradiction separation & stateless $-$ store arms $\ge$ 0.03 & \textbf{fail} (0.029)\\
C3 acknowledgment auditability & $= 100\%$ & pass (42/42)\\
C3 correction discrimination & false $\le$ true $-$ 0.10 & pass (0.420 vs.\ 0.875)\\
C4 provenance margin & no-prov.\ $-$ full $\ge$ 0.02 & \textbf{fail} (0.012)\\
Capability equivalence (CI criterion) & $\Delta$ accuracy CI $\subseteq \pm 0.05$ & pass\\
\bottomrule
\end{tabular}
\end{table}

\paragraph{What passed.}
The acknowledgment soundness guarantee and the correction-acceptance rule
hold on a real model. Every one of the 42 acknowledged revisions traces to an
accepted revision-log entry (42/42). This is the outcome the emission rule
guarantees by construction, so the check verifies implementation correctness
rather than discovering behavior. The check is on event linkage, not on
surface form: 31 of the 42 are correction turns that render the admission
template, and 11 are ask turns flagged for a retrieval conflict resolved at
response time, whose visible answer is an ordinary hedge or assertion. Ten of
the 31 rendered admissions retract a stored prior value the layer had not
itself uttered earlier in that conversation. The reciprocal direction is a
measured 21\% acknowledgment-flag incidence (42 of 197 accepted revisions
flagged). The 197 comprise 155 unconditionally accepted corrections of unheld
claims, 31 accepted corrections of held values, and 11 retrieval conflicts.
Every accepted change to a held value is flagged, and 31 of the 42 flags
render the admission utterance (31/197 = 15.7\% of all accepted revisions;
31/42 = 73.8\% of accepted changes to held values), while the 155
acquisitions are logged but not uttered.

The pre-specified correction check passes: the rule accepts 87.5\% of true
corrections and 42.0\% of false ones in Stage~B (105/120 and 81/193 scored
corrections; the frozen benchmark injects 422 per arm, 229 true and 193
false, and the scorer drops the 109 true corrections that agreed with the
stored value). That scored set, however, merges two paths of the rule.
Corrections of claims the store did not yet hold are accepted
unconditionally (93 of 93 true and 62 of 62 false), so the true-side rate is
dominated by acquisition, not revision. A post hoc split (not pre-specified;
results file \texttt{sensitivity\_post\_review.csv}) isolates the
corrections that contradicted a held belief: 12 of 27 true (0.444) versus 19
of 131 false (0.145). On held beliefs the gap (0.30) exceeds the 0.10 margin
on the point estimate. Its fact-clustered 95\% interval, $[0.09, 0.54]$
(resampling the 53 subjects with an eligible held correction), excludes zero
but reaches below the margin, so the contrast is positive and only the point
estimate clears the pre-specified margin. The absolute false-acceptance rate
on the scored set remains substantial (0.42--0.48 across grids), mostly
through the unheld path and partly through the acceptance rule's disclosed
weak-belief edge. The discrimination mechanism on held beliefs is the
corroboration clause operating over the benchmark's constructed retrieval
store (34 of 60 subjects present, one stale). True corrections to strongly
held beliefs are accepted at the rate at which that mostly-fresh store
corroborates them, while false corrections rarely find corroboration, so the
size of the gap scales with the constructed store's coverage and freshness
and is not a store-independent property of the layer. By the rule's
structure this gap is largely designed in, and a further stratification by
belief strength is not reported. The Stage-C scored rates are
0.871 vs.\ 0.435, non-overlapping under both conversation-clustered CIs
($[0.795, 0.939]$ vs.\ $[0.373, 0.497]$) and fact-clustered CIs
($[0.787, 0.948]$ vs.\ $[0.355, 0.533]$); the held-belief split there is 12
of 27 versus 20 of 129, with the same fact-level caveat.

Capability passes the specified CI-inclusion criterion and, separately, shows
a small directional improvement. The two statements are compatible because
equivalence bounds the magnitude while the CI locates its sign: when the full
layer answers, it is right 92.4\% of the time versus 90.6\% for uniform
($\Delta = +0.018$, 95\% CI $[0.006, 0.030]$, inside the $\pm 0.05$ margin
and excluding zero), because the expression policy filters the model's least
reliable content into declines. In the full arm the gating table's decline
cells already do this: switching the not-knowing floor off leaves every
Stage-B full-arm event unchanged, so the filtering is attributable to the
table rather than to the floor specifically. The price is coverage
(0.833 vs.\ 0.925). Both accuracies condition on each arm's own answered
subset, so the criterion certifies answer quality at the achieved coverage,
not unconditional task performance; a matched-coverage comparison is future
work. Confidently-asserted falsehoods drop from 0.087 (uniform) to 0.018
(full).

\paragraph{What failed.}
The C2 fidelity margin fails through a \emph{conditional degeneracy} of the
metric, an anchor--accuracy alignment: the uniform arm's expression-ECE is
near zero (0.0025--0.035) against the full layer's 0.25--0.27. The mechanism
is visible in the reliability bins (Figure~\ref{fig:reliability}): the
pipeline's asserted content (filtered by declines, upgraded by retrieval,
corrected by users) lands at roughly 0.9 accuracy overall, which is the
\textsc{assert} anchor; a degenerate policy that asserts everything
therefore scores as ``perfectly calibrated'' by anchor-based ECE whenever
corpus accuracy sits near one anchor, while the full layer's hedged bins are
punished for \emph{under}confidence: its \textsc{hedge-high} content is 96.8\%
accurate against a 0.6 anchor, largely because inferred and told content is
capped below \textsc{assert} by the gating table (the full arm's 660
\textsc{hedge-high} answers are 609 inferred and 51 told claims; retrieved
claims, at confidence 0.80, all assert). This is the design tension flagged
in Section~\ref{sec:stance}, now measured. The C3
contradiction margin fails for a scoping reason: under greedy decoding with
per-fact caching, the base model is deterministic, so the stateless arm
usually re-answers identically; its residual self-contradiction (0.029, just
under the 0.03 margin, against zero for all store arms) comes from stochastic
resolution-path flips (a re-ask that resolves by retrieval where the first
ask resolved parametrically), not from decoding noise. The store's measurable
value on a deterministic model shows up not in contradiction reduction but in
correction responsiveness, which the stateless arm lacks entirely. The C4
margin fails narrowly (0.0305 vs.\ 0.0182: a 0.0123 gap against a 0.02
margin): on this model most of the confidently-false reduction is attributable
to confidence gating with declines (shared by the no-provenance arm; the
three-state floor itself is redundant on the full arm's evaluated paths)
rather than to provenance typing as such. The acknowledgment-conditioning
effect (full vs.\ store-without-acknowledgment) is null on both stages'
contradiction measures. The null is structural,
because the contradiction scorer legitimizes a change through the accepted
log entry whether or not it is uttered and acknowledgment does not feed back
into the store, so this contrast cannot detect a behavioral effect of the
utterance. The two arms also draw independent retrieval-sampling streams and
differ on 239 of their 1{,}869 ask-turn outputs for that reason alone, which
is why every arm contrast in this paper is reported as a single realization
rather than a causal isolation.

\paragraph{Post hoc Amendment 1: an anchor-degeneracy-resistant check.}
Because the C2 failure is attributable to the metric's anchor dependence, we
added a protocol amendment after the Stage-B grid ran. The amended
manipulation check is expression-discrimination
AUC, the probability that a correct expressed claim carries a strictly
higher expression category than an incorrect one (ties one half). The
statistic is computed over expressed claims only (declines carry no
asserted content, are excluded here, and are counted in coverage), so each
arm's AUC conditions on its own expressed set and cross-arm comparisons span
the arms' achieved coverages. Any
single-category policy scores exactly 0.5 by construction, so the controls
are definitional baselines and the substantive bar is the 0.60 floor; the
anchor coincidence that broke ECE cannot recur, and the statistic matches the
framework's requirement (evidence-\emph{sensitive} expression) more directly
than anchor placement does. The amended margin (full $\ge 0.60$
and $\ge$ each control $+0.05$ under both extractors) was set against the
chance value and committed to the protocol file before the discrimination
statistic was computed on any arm. A coarse bin-based reading of the full arm's
combined reliability had been made during
the ECE failure analysis, but no per-extractor or per-arm AUC existed when the
margin was fixed. The original ECE verdicts above are unchanged. The Stage-B
configuration passes on the point estimate (combined AUC
0.653, 95\% CI $[0.598, 0.718]$, with the lower bound just below the 0.60
threshold; both controls exactly 0.5), and the consistency extractor alone
passes (0.661, 95\% CI $[0.604, 0.727]$, above chance). The layer gated on
the logit extractor alone, however, is \emph{anti-discriminative}: AUC 0.410, 95\% CI
$[0.366, 0.454]$, below chance under the pre-specified conversation unit
(under fact-level resampling the same estimate carries a 95\% interval of
$[0.30, 0.57]$, which includes chance, so the below-chance reading is
unresolved at the fact level). The inversion is a property of the gated
pipeline and is not observed under the threshold-only map: the same signal
there is weakly discriminative (0.569, 95\% CI $[0.543, 0.596]$; a
continuous raw-signal AUC was not computed). On this overconfident model the
pattern is consistent with high token probability lifting fluent
hallucinations into \textsc{assert} while the provenance caps hold accurate
inferred and told content below it; the per-extractor robustness
requirement therefore fails, and the amended check fails with it. Two further decompositions qualify the result. The threshold-only arm has a
higher AUC point estimate than the full layer (0.715 vs.\ 0.653 combined,
though their CIs overlap: $[0.681, 0.755]$ vs.\ $[0.598, 0.718]$), so the
provenance caps impose source-caution at an apparent discrimination cost, and
the Stage-A grid shows the same ordering in miniature (full 0.635
vs.\ controls 0.5). The failure is not a second metric
artifact; it is an extractor-dependence result. Gated end to end, the two
confidence sources yield oppositely signed discrimination, and a layer gated
on their conservative minimum
inherits the logit's rankings whenever that signal binds, which, measured
over the per-fact cache, is 30\% of facts (18 of 60 where the logit confidence is the
minimum). These per-arm AUC intervals are conversation-clustered; the
fact-level sensitivity analysis reported below covers the Stage-C selection
gates and, above, the logit-gated headline.

\paragraph{Stage C: the identified configuration under a separate commit.}
Stage~C evaluated the configuration the extractor-dependence finding points
to (the layer gated on the consistency extractor alone, with model, cache,
benchmark, arms, seeds, and margins otherwise identical) and committed that
freeze to the repository before the Stage-C grid ran, so the ordering is
attested by commit separation. The freeze's prior-knowledge register is
explicit: the fidelity-arm discrimination values under consistency gating
were already known from Stage~B's robustness runs (full 0.661, controls
exactly 0.5, so the manipulation check was expected to pass); every
non-fidelity-arm outcome was unknown, and the downstream selection decision
was specified to hinge on capability
equivalence under consistency-only gating. The results: the manipulation check, computed like every discrimination
number in this paper over the layer's \emph{emitted expression categories}
and never over the raw extractor signal, so that it tests the expression
layer end to end, passes as expected
(AUC 0.661, 95\% CI $[0.604, 0.727]$, margin $\ge 0.60$ and $\ge$ controls
$+0.05$); capability equivalence, the outcome that was unknown when the protocol was
frozen, passes ($\Delta = +0.028$, 95\% CI $[0.016, 0.042]$,
within $\pm 0.05$ and excluding zero); acknowledgment auditability is
again 42/42 and the acceptance rule again discriminates (0.871 vs.\ 0.435).
The C3 and C4 margins fail exactly as in Stage~B (stateless contradiction
0.023, 95\% CI $[0.004, 0.048]$, below the 0.03 bar; provenance gap 0.009
against 0.02), so those findings stand unchanged, and we claim nothing from contradiction
reduction on this model class: C3's supported
content is correction-acceptance behavior (the discrimination above) and
acknowledgment auditability. The
threshold-only arm still carries the higher discrimination point estimate
(0.730, 95\% CI $[0.697, 0.770]$, overlapping full's), so we still claim no
statistically resolved provenance-cost difference.

Because Stage~C re-scores the same frozen instance set that informed the
configuration choice, its confirmatory weight on an independent draw could be
questioned. A robustness draw, a freshly seeded regeneration of the benchmark (new
instances, new injections; generator, world, model, arms, and margins
identical) frozen in the protocol before it ran, replicates the
conversation-level criteria: manipulation check
0.677, 95\% CI $[0.635, 0.726]$; capability equivalence $\Delta = +0.023$,
95\% CI $[0.009, 0.036]$; auditability 56/56; correction discrimination
0.892 vs.\ 0.482. On precision: with 120 conversations per arm the achieved
cluster-bootstrap 95\% half-widths on rates run roughly 0.01--0.03 (the
intervals reported throughout), so the narrowly failed margins are
point-estimate failures that the achieved precision leaves statistically
unresolved (the stateless-contradiction CI $[0.004, 0.048]$ spans the 0.03
margin, and no paired interval was computed for the C4 gap); no prospective
power analysis was conducted. A further conditioning caveat: every grid,
including the robustness draw, reads the same per-fact model cache, so all
intervals condition on a single realization of the $k{=}8$ consistency
samples per fact; extractor-sampling variability is not propagated, and the
Stage-C manipulation bound's proximity to the 0.60 bar (lower limit 0.604)
should be read with that conditioning in mind. A further sensitivity analysis clusters
uncertainty at the \emph{fact} level instead: because the per-fact model
cache makes all 120 conversations recycle the same 60 facts, conversation
clustering (the unit specified for Stage~B) understates fact-level
dependence. Under subject-level resampling the point estimates are unchanged
but the intervals widen: manipulation AUC 0.661, fact-clustered
95\% CI $[0.452, 0.885]$ (including chance); equivalence
$\Delta = +0.028$, $[+0.006, +0.057]$ (exceeding the $+0.05$ bound); the
fresh draw behaves alike (0.677, $[0.480, 0.862]$; $+0.023$,
$[-0.006, +0.057]$). Sixty distinct facts are simply too few to resolve the
gates at the fact level, and the robustness draw, which redraws
conversations rather than the world, mitigates instance-level overfitting but not
generator-distribution artifacts. Applying the same fact-level clustering to
the scored-set correction-discrimination result leaves it intact (CIs above),
but the held-belief margin and the logit-gated below-chance AUC both lose
resolution at the fact level (Section~\ref{sec:results}; the held-belief
contrast itself stays positive); of the supported
results, only the audit guarantee and the scored-set correction contrast are
insensitive to the clustering unit.

\clearpage

\section{Ablation handoff and evidential scope}
\label{sec:handoff}

The downstream perception study requires a truth-present versus truth-ablated
pair only if expression is evidence-sensitive and answer-when-given accuracy
meets the equivalence criterion. Stage~B satisfies capability equivalence but
fails both the original ECE criterion and the post hoc AUC criterion because
expression gated on the logit extractor is anti-discriminative on this model. Stage~B therefore
does not provide a qualifying condition pair.

Stage~C evaluates consistency-only gating as a data-dependent
reconfiguration under a separately committed protocol with a prior-knowledge
disclosure. Its manipulation AUC was already expected from the Stage-B
per-extractor analysis; capability equivalence was unknown. Both
conversation-level criteria pass and the result repeats on a redrawn set of
conversations over the same fact base (Section~\ref{sec:results}). The
corresponding configurations and outputs constitute a candidate condition
pair for downstream validation. They do not establish that the manipulation
is ready for a human study: fact-clustered intervals leave both criteria
unresolved at 60 facts, so the fact base must be expanded first. The original
ECE failures remain unchanged, and the selected configuration's AUC is not
independent confirmation because it motivated the selection.

The evidence supports acknowledgment soundness (a by-construction guarantee
verified in all audited events, 42/42 in the primary real-model grid) and
discrimination between true and false corrections of held beliefs (0.44
vs.\ 0.15 in Stage~B, a post hoc split whose fact-clustered interval excludes
zero while only its point estimate clears the 0.10 margin; the pre-specified
scored rates of 0.875 vs.\ 0.420 are dominated by the unconditional
acceptance of corrections to unheld claims). The latter is an effect of
retrieval corroboration whose size depends on the constructed store's
coverage and freshness, and 42--48\% of scored false corrections are still
accepted across grids, most through the unheld path and the rest under the
disclosed weak-belief rule. It does not show that the three-state or provenance mechanisms improve
expression discrimination relative to the threshold-only arm. Consequently,
any later perception effect from this pair would identify
evidence-congruent hedging in general, not those mechanisms specifically. The
scope is one 0.5B model, one 60-fact world, and a consistency extractor; other
models require separate evaluation.

\section{Limitations and scope}
\label{sec:honesty}

\emph{Expressed epistemic behavior, not epistemology}: the layer is designed to
produce calibrated expression, not knowledge. On the real model its verbal
calibration failed the pre-specified checks (Section~\ref{sec:results}) while
its filtering worked; when the base model is wrong, the layer expresses the
wrong answer just as fluently, and the confidently-false rates quantify this
residue. \emph{Confidence estimation and expression remain distinct}:
confidence estimation is imported
\cite{farquhar2024detecting,kadavath2022language}; the contribution is the
behavior layer; fidelity is the manipulation check, never the stance's
definition. \emph{Provenance is instrumentation}: a pipeline-path tag with an
acknowledged ambiguous boundary, not an epistemological taxonomy. \emph{The
benchmark is constructed}: contradictions and false corrections are injected
so scoring is mechanical against known ground truth; this measures the
operator on knowable cases, not open-domain consistency, and no model judge
appears anywhere in the loop. \emph{One small model, one machine}: Stage-B
claims are scoped to the 0.5B model class tested; the deterministic-decoding
scoping of C3 and the sign-inverting logit extractor are part of that
scope (the unnormalized mean-probability feature leaves length bias as a
rival explanation the instrument can probe, and a sampled-decoding
replication of the contradiction metric is future
work); larger or better-calibrated models may place the extractor-dependence
boundary elsewhere, which the included instrument can measure but this paper
does not claim. \emph{Introspective reports are bounded by the role-play
frame}~\cite{shanahan2023roleplay}: no claim of access to real internal
states. \emph{No trust measurement}: whether expressed doubt,
visible revision, and source-aware caution change perceived trustworthiness
or mind-likeness is the deferred human-subjects question; nothing here
licenses that conclusion. \emph{The title names the design target}: on the
tested model the demonstrated results are the audit guarantee, the
correction-discrimination contrast, and the negative gating findings;
whether the engineered stance serves believability is the deferred
question.

\section{Series context}
\label{sec:series}

This paper is the third in a research program on believability as dimensional
completeness rather than capability~\cite{cochinescu2026perceivedagi}. It
operationalizes the program's truth dimension alongside companion work on
temporal continuity and on structured variation~\cite{cochinescu2026entropy}. A future preregistered
perception study is intended to test the resulting ablation pairs. These
series links provide motivation; the methods and claims evaluated here are
self-contained.

\paragraph{Data and code availability.}
{\sloppy
The layer (\texttt{truthllm}), the versioned benchmark (generator, frozen
instances, SHA-256 manifest, language-agnostic scorer), the frozen protocol
with both stages' pre-specified margins and the Amendment~1 disclosure, all
results files behind every number in this paper, and the one-command
reproduction scripts (\texttt{scripts/reproduce.py},
\texttt{scripts/reproduce\_stageb.py}) are included in the public companion
repository: \url{https://github.com/cochinescu/truth-llm-prototype}. A full
project snapshot including the manuscript source, as of July 2026 (it
predates the September 2026 textual revisions and the post hoc diagnostics
file; every frozen number is unchanged), is archived at
\href{https://doi.org/10.5281/zenodo.21462986}{doi:10.5281/zenodo.21462986}
(MIT license), and the benchmark is archived separately as an independently
citable dataset at
\href{https://doi.org/10.5281/zenodo.21462988}{doi:10.5281/zenodo.21462988}
(CC~BY~4.0). The cached model outputs
and run metadata record the exact model revision and seeds. No human-subjects
or personal data were collected.\par}

\section*{Acknowledgments and tool-use disclosure}
Large language models were used as drafting and revision aids. They were not
used to label evaluation data, judge model outputs, or compute reported
statistics. The author directed and reviewed their use and takes
responsibility for the manuscript.

\bibliographystyle{plain}
\bibliography{Truth}

\end{document}